\documentclass[letterpaper, 10 pt, conference]{ieeeconf}  

\IEEEoverridecommandlockouts                              

\usepackage{graphicx} 
\usepackage{amsmath} 
\usepackage{amssymb}  

\usepackage{amsthm}
\usepackage{cite}

\usepackage{accents}
\usepackage{centernot}
\usepackage{color}
\usepackage{algorithm}
\usepackage{algorithmicx}
\usepackage[noend]{algpseudocode}
\usepackage[font=scriptsize]{caption}
\usepackage{subcaption} 
\usepackage{mathtools}
\usepackage{physics}

\usepackage{color}
\usepackage{soul}
\definecolor{mypurple}{rgb}{0.75,0.0,0.75}
\definecolor{myblue}{rgb}{0,0.4,0.4}
\def\*#1{\mathbf{#1}}
\usepackage{hyperref}
\renewcommand{\baselinestretch}{0.8928}     
\newcommand\fomref[1]{{#1^f_{\text{ref}}}}
\newcommand\romref[1]{{#1_{\text{ref}}}}

\usepackage{array}
\makeatletter
\newcommand{\thickhline}{%
    \noalign {\ifnum 0=`}\fi \hrule height 1pt
    \futurelet \reserved@a \@xhline
}
\newcolumntype{"}{@{\hskip\tabcolsep\vrule width 1pt\hskip\tabcolsep}}
\makeatother

\title{\LARGE \bf
Soft Robot Optimal Control Via
Reduced Order Finite Element Models 
}

\author{Sander Tonkens$^{1}$, Joseph Lorenzetti$^{2}$, and Marco Pavone$^{2}$
\thanks{$^{1}$Sander Tonkens is with the Department of Mechanical Engineering, Stanford University, Stanford, CA 94305, USA
        {\tt\small tonkens@stanford.edu}}%
\thanks{$^{2}$Joseph Lorenzetti and Marco Pavone are with the Department of Aeronautics and Astronautics, Stanford University, Stanford, CA 94305, USA
        {\tt\small \{jlorenze, pavone\}@stanford.edu}}%
\thanks{This work was supported by the Office of Naval Research (ONR) (Grant N00014-17-1-2749). Joseph Lorenzetti is supported by the Department of Defense (DoD) through the National Defense Science and Engineering Fellowship (NDSEG) Program.}
}

\begin{document}

\maketitle
\thispagestyle{empty}
\pagestyle{empty}


\begin{abstract}
Finite element methods have been successfully used to develop physics-based models of soft robots that capture the nonlinear dynamic behavior induced by continuous deformation.
These high-fidelity models are therefore ideal for designing controllers for complex dynamic tasks such as trajectory optimization and trajectory tracking. However, finite element models are also typically very high-dimensional, which makes real-time control challenging. In this work we propose an approach for finite element model-based control of soft robots that leverages model order reduction techniques to significantly increase computational efficiency. In particular, a constrained optimal control problem is formulated based on a nonlinear reduced order finite element model and is solved via sequential convex programming. This approach is demonstrated through simulation of a cable-driven soft robot for a constrained trajectory tracking task, where a 9768-dimensional finite element model is used for controller design.
\end{abstract}

\section{Introduction}\label{sec:intro}
Soft robots are an emerging class of robots that leverage natural compliance through continuous deformation to simplify tasks such as object manipulation, moving in complex environments, safely interacting with humans, and even assisting in surgical procedures \cite{RusTolley2015, ThuruthelAnsariEtAl2018}.
However, significant challenges in modeling, simulation, and control have limited their practical use. One fundamental challenge is that continuously deforming soft robots are infinite-dimensional systems that exhibit significant nonlinear behavior during structural deformation. Another is that diversity among soft robot designs makes it challenging to develop modeling techniques that are generalizable.

One approach to soft robot modeling and control is to hand-engineer simplified models of the robot's motion by making approximations.
For example piecewise constant curvature models \cite{WebsterJones2010, DellaSantinaKatzschmannEtAl2020}, beam models \cite{GravagneRahnEtAl2003}, and Cosserat models \cite{RendaBoyerEtAl2018} have been used. However, these low-fidelity approximations are typically tailored to specific robot geometries and often model only the robot's kinematics, making it challenging to design controllers for dynamic tasks.

Data-driven methods have also been developed to generate models directly from input-output data. This approach has been used to develop both kinematic and dynamic controllers. For example, \cite{BernSchniderEtAl2020} learns a differentiable kinematics model for solving inverse kinematics problems, and \cite{GillespieBestEtAl2018, ThuruthelFaloticoEtAl2019} learn neural network dynamics models to develop closed-loop controllers. Another data-driven approach is proposed in \cite{BruderGillespieEtAl2019}, which uses Koopman operator theory to build a dynamics model that is used for model predictive control. 
While data-driven methods are generalizable to different types of soft robots, they fail to leverage physics-based principles and there is no systematic procedure for developing them. 
Additionally, reliance on experimental data for model identification and controller validation precludes the use of data-driven modeling approaches in the design process.

Alternatively, finite element methods (FEMs) provide a systematic, physics-based approach to soft robot modeling. These approaches can be used to develop high-fidelity models for a wide variety of soft robots and can be directly incorporated into the design process. While some techniques directly use FEM models for inverse kinematics based control \cite{Duriez2013} or trajectory optimization \cite{BernBanzetEtAl2019}, the high-dimensionality of FEM models (e.g. thousands to tens of thousands of degrees of freedom) makes the design of real-time dynamic controllers challenging. 

In this work, we propose an approach that leverages high-fidelity finite element models to control cable-actuated soft robots. The challenge of computational efficiency is addressed by using model order reduction techniques to compress the high-dimensional FEM models without significant loss in modeling accuracy. The resulting reduced order models (ROMs) are then used to efficiently solve optimal control problems, such as trajectory tracking tasks.

\begin{figure}[t]
    \centering
    \includegraphics[width=0.7\columnwidth]{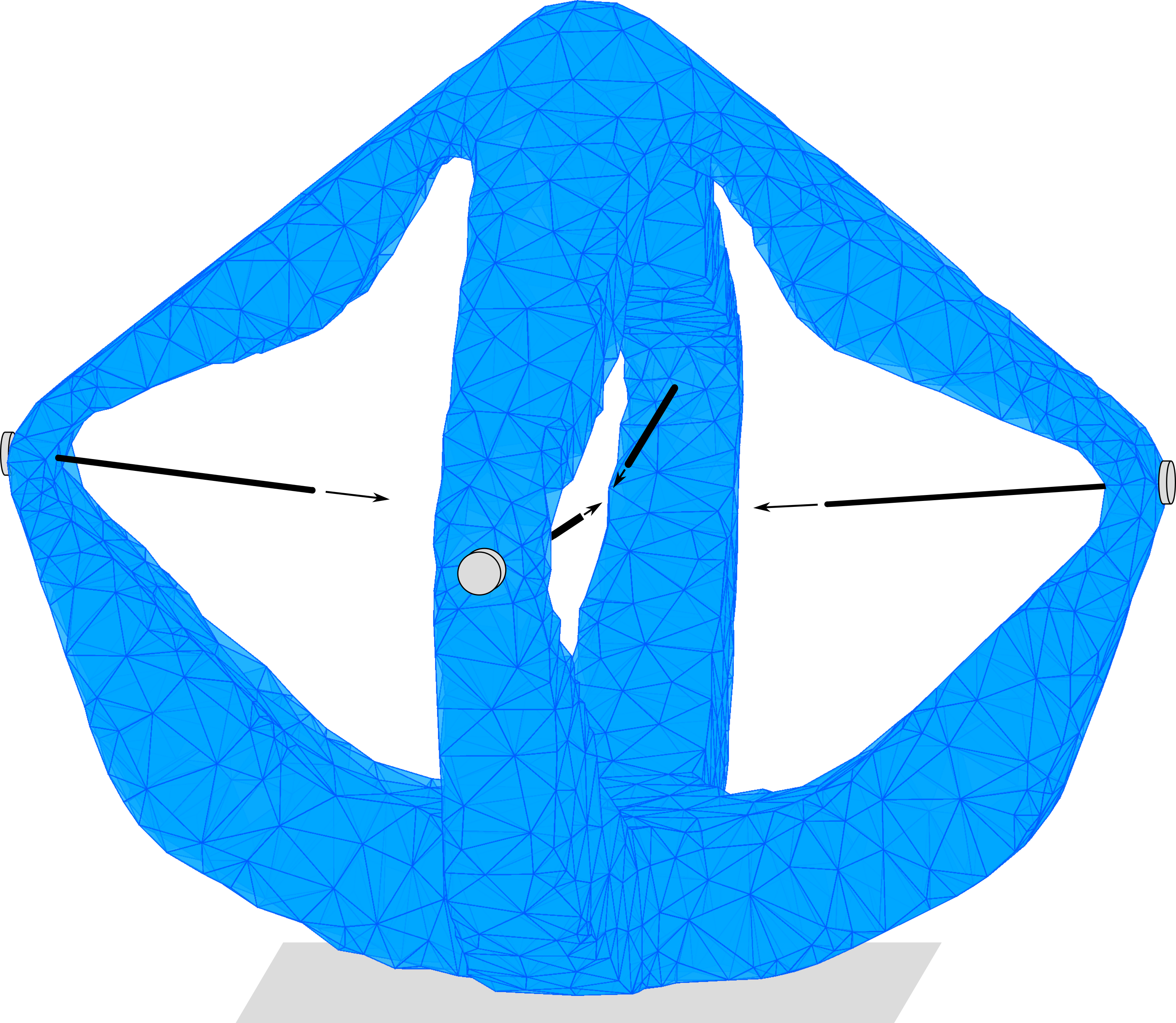}
    \caption{Finite element mesh for a ``Diamond'' soft silicone robot with $\mathcal{N} = 1628$ mesh nodes. This robot is fixed at the base and is actuated by controlling the tension in four cables (black) attached to the robot's ``elbows''. 
    Finite element methods offer a systematic approach to generating high-fidelity dynamics models of soft robots.
    }
    \label{fig:robots_fem}
\end{figure}

{\em Related Work:} A variety of principled model order reduction techniques have been developed for compressing high-dimensional dynamics models \cite{Antoulas2005}. 
In the context of soft robotics, these methods can be useful for reducing the dimensionality of FEM models by orders of magnitude without significant loss in accuracy.

Specifically, linearized reduced order FEM models have been used to design regulating output-feedback controllers for cable-actuated \cite{ThieffryKruszewskiEtAl2019} and pneumatically-actuated \cite{KatzschmannThieffryEtAl2019} soft robots, as well as trajectory tracking controllers \cite{ThieffryKruszewskiEtAl2019b}.
Linear ROMs have also been used in contexts beyond soft robotics for solving constrained optimal control problems \cite{LorenzettiPavone2020b, LorenzettiMcClellanEtAl2020}. However, using a \textit{linearized} FEM model is not sufficient for many dynamic control problems due to the significant nonlinearities arising from the soft robot's deformation. 
While \textit{nonlinear} FEM model reduction techniques exist, they have generally been developed for simulation applications and are difficult to use for controller synthesis. 
In fact, in the context of soft robotics, \textit{nonlinear} FEM model reduction has only been used to design inverse kinematics-based controllers \cite{GouryDuriez2018}, which are not sufficient for dynamic control tasks such as trajectory tracking.  
In summary, due to the use of either \textit{linearized} models or kinematics-based controllers, existing approaches that leverage reduced order FEM models cannot adequately address dynamic soft robot control tasks.

{\em Statement of Contributions:}
This work proposes an approach for constrained optimal control of soft robots that leverages high-fidelity \textit{nonlinear} finite element models.
We handle the computational challenges induced by the high-dimensionality of the FEM models by exploiting existing model order reduction techniques in the design of our control scheme.
In particular, we approximate the FEM model with a low-dimensional piecewise affine model that is amenable to optimal control. We design an output feedback receding horizon control scheme based on sequential convex programming, which can be used for a variety of dynamic control tasks including setpoint and trajectory tracking.
We demonstrate the performance of the proposed controller in simulation, using the soft robot shown in Figure \ref{fig:robots_fem} with a finite element model of dimension 9768.
An implementation of the approach\footnote{\href{https://github.com/StanfordASL/soft-robot-control}{github.com/StanfordASL/soft-robot-control}} that leverages SOFA \cite{CoevoetMorales-BiezeEtAl2017}, an open-source FEM software toolkit, is also provided.

{\em Organization:} We begin by presenting a general form of the nonlinear soft robot FEM model and the associated optimal control problem in Section \ref{sec:prob}. Next, in Section \ref{sec:rom} we use reduced order modeling techniques to construct a ROM, which is then used for computationally efficient optimal control in Section \ref{sec:control_scheme}. We present simulation results in Section \ref{sec:results} and conclude with the merits of our approach and avenues of future work in Section \ref{sec:conclusion}.

\section{Problem Formulation}\label{sec:prob}
We begin by defining the soft robot finite element model and formulating the control problem.

\subsection{Soft Robot Finite Element Model}\label{sec:fom_definition}
Finite element methods are a numerical approach for solving partial differential equations (PDEs) and have been used for physics-based modeling and simulation in many domains, including fluid and structural mechanics as well as soft robotics \cite{CoevoetMorales-BiezeEtAl2017}.
These methods solve PDEs by performing a spatial discretization at a finite number of nodes defined by a mesh, which results in a finite set of ordinary differential equations corresponding to the state of each mesh node. 

In the context of soft robotics, the state of each node consists of the node's position and velocity, resulting in six degrees of freedom. Therefore, for a spatial discretization with $\mathcal{N}$ nodes the resulting FEM model will have a total dimension of $n^f = 6\mathcal{N}$ (FEM model variables are denoted using $(\cdot)^f$). In particular, the soft robot FEM model is defined as in \cite{GouryDuriez2018} using Newton's second law:
\begin{equation}\label{eq:fom_dynamics}
\begin{split}
    M^f\dot{v}^f &= P^f - F^f(q^f, v^f) + H^f(q^f)u, \\
    \dot{q}^f &= v^f,
\end{split}
\end{equation}
where $q^f(t) \in \mathbb{R}^{3\mathcal{N}}$ is the vector of node positions, $v^f(t) \in \mathbb{R}^{3\mathcal{N}}$ is the vector of node velocities, and $u(t) \in \mathbb{R}^m$ is the control input. Additionally, $M^f \in \mathbb{R}^{3\mathcal{N} \times 3\mathcal{N}}$ is a constant mass matrix, $P^f \in \mathbb{R}^{3\mathcal{N}}$ represents constant external forces (e.g. gravity), $F^f(q^f, v^f): \mathbb{R}^{3\mathcal{N}} \times \mathbb{R}^{3\mathcal{N}} \rightarrow \mathbb{R}^{3\mathcal{N}}$ is a nonlinear function that defines the internal forces (e.g. stresses from deformation), and $H^f(q^f):\mathbb{R}^{3\mathcal{N}} \rightarrow \mathbb{R}^{3\mathcal{N}\times m}$ is a nonlinear function that specifies the input matrix.
The internal forces $F^f(q^f, v^f)$ can be modeled in several ways depending on the material properties of the robot.
A classic example is to use a linear-elastic model, which assumes a linear stress-strain relationship through Hooke's law.

To enable output feedback control we assume a linear measurement model given by:
\begin{equation}\label{eq:measurement}
\begin{split}
    y &= C_y^f x^f,
\end{split}
\end{equation}
where $x^f = [{v^f}^T, {q^f}^T]^T \in \mathbb{R}^{n^f}$ is the combined state vector and $y(t) \in \mathbb{R}^p$ is the measurement. Additionally, in FEM-based control problems only a small set of variables may be of particular interest (e.g. end-effector position and velocity). We denote these performance variables as $z(t) \in \mathbb{R}^o$ and assume a linear performance output model:
\begin{equation}\label{eq:performance}
\begin{split}
    z &= C_z^f x^f.
\end{split}
\end{equation}
This output model can be used to simplify the formulation of the optimal control problem by expressing the cost function and constraints in terms of the relevant performance variables rather than the full state $x^f$ of the FEM model.

\subsection{Constrained Optimal Control Problem}\label{sec:fom_ocp}
In this section we introduce the soft robot optimal control problem (OCP). First, a set of constraints on the performance variables $z$ and control $u$ are assumed to be defined by:
\begin{equation}\label{eq:constraints}
    u \in \mathcal{U}, \hspace{1 cm} z \in \mathcal{Z}, 
\end{equation}
where $\mathcal{U} \vcentcolon=\{u \mid H_u u \le b_u \}$ and $\mathcal{Z} \vcentcolon=\{ z \mid H_z z \le b_z \}$ are convex polytopes with $H_u \in \mathbb{R}^{n_u \times m}$ and $H_z \in \mathbb{R}^{n_z \times o}$. Second, a cost function is defined over a finite horizon as:
\begin{equation}\label{eq:objective_fom}
\begin{split}
    J^f =\lVert \delta z(t_f)\rVert_{Q_f}^2 + \int\limits_{t_0}^{t_f} \lVert \delta u(t) \rVert_{R}^2 + \lVert \delta z(t) \rVert_{Q}^2 dt,
\end{split}
\end{equation}
where $[t_0, t_f]$ defines the time horizon, $Q, Q_f \in \mathbb{R}^{o \times o}$ are positive semi-definite performance variable cost matrices and $R \in \mathbb{R}^{m \times m}$ is a positive definite control cost matrix. The terms $\delta z(t) = z(t) - z_d(t)$ represent deviations of the performance variables with respect to a desired target trajectory $z_d(t)$, and $\delta u(t) = u(t) - u_d(t)$ represents deviations from a desired target input $u_d(t)$. For example, in trajectory tracking tasks $z_d(t)$ defines the reference trajectory (and typically $u_d(t) = 0$), and in regulation tasks $z_d(t) = z_0$, $u_d(t) = u_0$ where $z_0$ and $u_0$ are the desired equilibrium values.

The FEM-based soft robot constrained optimal control problem is then formulated as:
\begin{equation}\label{eq:ocp}
    \begin{split}
        \underset{q^f,v^f,u}{\text{minimize}} \quad &J^f, \\
        \text{subject to}\quad &         M^f\dot{v}^f = P^f - F^f(q^f, v^f) + H^f(q^f)u, \\
    &\dot{q}^f = v^f, \\
& u \in \mathcal{U}, \quad z \in \mathcal{Z}, \quad z = C^f_z \begin{bmatrix}
        v^f \\ q^f\end{bmatrix}.
    \end{split}
\end{equation}
A common approach for solving these types of optimal control problems is to transcribe the finite-horizon OCP into a nonlinear programming problem that can be solved with existing numerical algorithms.
This problem can then be used for closed-loop feedback control by repeatedly solving the optimization problem in a receding horizon fashion.

However, practical implementations of receding horizon control schemes require the OCP to be solved in real-time, which is challenging in this context for several reasons.
First, the FEM model \eqref{eq:fom_dynamics} is nonlinear and therefore the OCP contains non-convex constraints. Second, even the simple case where the optimization problem is a convex quadratic program has a computational complexity that is $\mathcal{O}(T(\mathcal{N} + m)^3)$, where $T$ is the number of time steps in the OCP horizon \cite{WangBoyd2010}. 
This generally precludes the direct use of FEM models for optimization-based control of soft robots since fine spatial discretizations (a large number of nodes $\mathcal{N}$) are required for high-fidelity modeling. In fact, it is not uncommon for the full state dimension $n^f$ to be on the order of thousands to tens of thousands, making it impossible to solve the OCP in real-time with existing state-of-the-art optimization algorithms.

Our proposed approach to address this computational challenge is to leverage model order reduction techniques to derive a high-fidelity (but low-dimensional) approximation to the original soft robot FEM model. This reduced order model can then be directly used for real-time soft robot control.

\section{Reduced Order Modeling for Soft Robots}\label{sec:rom}
Principled model order reduction techniques have been developed to derive high-fidelity reduced order models (ROMs) from high-dimensional models. For nonlinear systems, the model order reduction process generally consists of two successive steps. The first is to project the high-dimensional model onto a reduced-order subspace, and the second is to define an efficient approach for approximately evaluating the model's high-dimensional nonlinear terms.

\subsection{Projection-based Model Order Reduction}\label{sec:projection_mor}
In this work we consider projection-based methods, which are a general class of model order reduction methods that project the high-dimensional model onto a reduced order subspace and include widely-used approaches such as balanced truncation and proper orthogonal decomposition (POD) \cite{Antoulas2005}. 
In particular, we use POD to define a Galerkin projection specified by a projection matrix $UU^T$, where $U \in \mathbb{R}^{3\mathcal{N} \times r}$ is an orthogonal basis matrix and $r$ is the dimension of the reduced-order subspace (typically $r \ll 3\mathcal{N}$). This projection is applied to the high-dimensional FEM position and velocity vectors to define reduced order position and velocity vectors:
\begin{equation}\label{eq:project}
q=U^T(q^f - \fomref{q}), \quad v=U^T(v^f - \fomref{v}),
\end{equation}
and to approximately reconstruct the original states by:
\begin{equation}\label{eq:reproject}
q^f\approx Uq + \fomref{q}, \quad v^f\approx Uv + \fomref{v},
\end{equation}
where $\fomref{q}, \fomref{v} \in \mathbb{R}^{3\mathcal{N}}$ are constant reference states that are used to better condition the basis matrix $U$. In particular, we select $\fomref{q}$ and $\fomref{v}$ to correspond to a static equilibrium (i.e. $\fomref{v} = 0$) of the soft robot.

The reduced order model is then defined by projecting the FEM model \eqref{eq:fom_dynamics} onto the reduced order subspace:
\begin{equation}\label{eq:projection_rom}
    \begin{split}
        M \dot{v} &= P - U^T F^f(Uq + \fomref{q}, \:Uv + \fomref{v}) \\
        & \quad + U^T H^f(Uq + \fomref{q})u, \\
        \dot{q} &= v + \romref{v},
    \end{split}
\end{equation}
with $M = U^T M^f U$, $P = U^T P^f$, and $\romref{v} = U^T\fomref{v}$. Additionally, with the combined reduced order state $x = [{v}^T, {q}^T]^T \in \mathbb{R}^{n}$ (where $n=2r$) and reference state $\fomref{x} = [\fomref{v}^T, \fomref{q}^T]^T \in \mathbb{R}^{n}$, the reduced order measurement and performance models are given by:
\begin{equation}
y = C_y x + \romref{y}, \quad z = C_z x + \romref{z},
\end{equation}
where $C_y = C_y^f V$ and $C_z = C_z^f V $ with $V$ defined as $V=\text{blkdiag}(U, U)$, and $\romref{y} = C_y^f \fomref{x}$ and $\romref{z} = C_z^f\fomref{x}$ are constants.
Crucially, this projection results in a ROM \eqref{eq:projection_rom} with combined dimension $n$, which can be orders of magnitude smaller than the original FEM model \eqref{eq:fom_dynamics} of dimension $n^f$.

\subsection{Nonlinear Model Reduction}\label{sec:nonlinear_mor}
While the ROM \eqref{eq:projection_rom} has a reduced dimension, evaluation of the nonlinear terms is still computationally expensive because of their dependence on the high-dimensional state (e.g. the evaluation of the internal force $U^T F^f(Uq + \fomref{q}, \:Uv + \fomref{v})$ scales as $\mathcal{O}(n^f)$). In this work we address this by approximating the nonlinear terms by reduced order piecewise affine functions, resulting in a piecewise affine ROM. 

Specifically, the piecewise affine approximations are generated by considering $N$ linearization points $(q_i, v_i)$, which are represented in the high-dimensional space by $q^f_i = Uq_i + \fomref{q}$ and $v^f_i = Uv_i + \fomref{v}$.
Around each of these linearization points, the internal forces $F^f$ are approximated by a first-order Taylor expansion:
\begin{equation*}
    \begin{split}
        F^f(q^f, v^f) \approx & F^f\rvert_{q^f_i, v^f_i} + \frac{\partial{F^f}}{\partial q^f}\rvert_{q^f_i, v^f_i}(q^f - q^f_i) \\ &+ \frac{\partial{F^f}}{\partial v^f}\rvert_{q^f_i, v^f_i}(v^f - v^f_i).
    \end{split}
\end{equation*}
The internal force terms from \eqref{eq:projection_rom} can then be written as:
\begin{equation*}
     \begin{split}
        U^T F^f(Uq + \fomref{q},&\: Uv + \fomref{v}) \approx \\
        &F_i + K_i(q - q_i) + D_i(v-v_i),
    \end{split}   
\end{equation*}
with reduced order terms:
\begin{equation*}
\begin{split}
    F_i &= U^TF^f\rvert_{q^f_i, v^f_i}, \quad K_i = U^T \frac{\partial{F^f}}{\partial q^f}\rvert_{q^f_i, v^f_i}U, \\
    D_i &= U^T \frac{\partial{F^f}}{\partial v^f}\rvert_{q^f_i, v^f_i}U. \\
\end{split}
\end{equation*}
The input matrix $H^f$ in \eqref{eq:projection_rom} is then simply approximated by the zeroth-order Taylor series expansion:
\begin{equation*}
H^f(Uq + \fomref{q}) \approx H^f(q^f_i).
\end{equation*}
With these simplifications the ROM \eqref{eq:projection_rom} can be approximated in the vicinity of $x_i = [v_i^T, q_i^T]^T$ by:
\begin{equation}\label{eq:single_lin}
\dot{x} = A_ix + B_iu + d_i,
\end{equation}
where:
\begin{equation*}
\begin{split}
A_i &= \begin{bmatrix}-M^{-1}D_i & -M^{-1}K_i \\ I & 0\end{bmatrix}, \quad B_i = \begin{bmatrix} M^{-1}H_i \\ 0 \end{bmatrix}, \\
d_i&= \begin{bmatrix}M^{-1}(P - F_i + K_i q_i + D_i v_i) \\ \romref{v} \end{bmatrix}.
\end{split}
\end{equation*}

Combined, these $N$ linearization points provide a global approximation of \eqref{eq:projection_rom} via the piecewise affine ROM:
\begin{equation}\label{eq:dynamics_rom}
\begin{split}
\dot{x} &= \begin{cases}
A_ix + B_iu + d_i, \quad i = \arg\min_j \lVert x - x_j \rVert_W,
\end{cases}
\end{split}
\raisetag{1\normalbaselineskip}
\end{equation}
where $W\in\mathbb{R}^{n\times n}$ is a positive semi-definite weighting matrix that defines a distance metric for determining the nearest linearization point. For example, we choose $W=\text{blkdiag}(0, I)$ for the soft robot in Figure \ref{fig:robots_fem} since the internal forces are heavily dependent on the robot's configuration $q^f$.

Since the matrices $A_i$, $B_i$, and $d_i$ are of reduced order, the ROM \eqref{eq:dynamics_rom} can easily be stored in memory, efficiently used for simulation, and its Jacobians can be trivially extracted for use in optimization-based control.
A discrete-time version of this model can be defined by:
\begin{equation}\label{eq:discrete_rom}
\begin{split}
x_{k+1} = g(x_k, u_k) \coloneqq &\begin{cases}
A_{i,d}x_k + B_{i,d}u_k + d_{i,d},
\end{cases}
\\
& \quad \quad i = \arg\min_j \lVert x_k - x_j \rVert_W,
\end{split}
\end{equation}
where $A_{i,d}, B_{i,d}$ and $d_{i,d}$ are the discretized reduced order matrices, which can also be computed \textit{a priori}.

Not only is the use of piecewise affine models common for nonlinear control applications, it is also a popular approach in the model order reduction community and is known as the trajectory piecewise linear (TPWL) method \cite{RewienskiWhite2003}.

\subsection{Building Piecewise Affine ROM}\label{sec:building_ROM}
This section proposes an automated approach for the development of the piecewise affine ROM \eqref{eq:dynamics_rom}, specifically the computation of the reduced order basis matrix $U$ and the selection of the linearization points $(q_i, v_i)$. 
This process relies on simulations of the high-dimensional FEM model \eqref{eq:fom_dynamics}, but can be done entirely \textit{offline}.

As mentioned in Section \ref{sec:projection_mor}, the basis matrix $U$ that defines the reduced order subspace is computed via POD, a data-driven method for compressing high-dimensional physics models. 
In this work, POD is implemented by simulating the FEM model \eqref{eq:fom_dynamics} to collect a set of ``snapshots'', which can include the soft robot's configuration $q^f$, velocity $v^f$, and acceleration $\dot{v}^f$, and implicitly defines a basis that characterizes the robot's behavior. A principal component analysis of the snapshots is then used to identify the reduced order subspace. In particular, the subspace is selected by defining a ``snapshot matrix'' $S$ with columns corresponding to snapshots, and then computing a singular value decomposition $S = \bar{U} \bar{\Sigma} \bar{V}$. The basis matrix $U$ is then defined by taking the $r$ columns of $\bar{U}$ associated with the $r$ largest singular values.
The dimension of the subspace, $r$, is typically chosen to be as small as possible while still providing good approximation accuracy. In practice, a commonly used heuristic for quantifying the approximation accuracy is the ``energy'' of the truncated singular values \cite{Antoulas2005}.

The linearization points used to define the piecewise affine ROM \eqref{eq:dynamics_rom} are also determined via an \textit{offline} data-driven procedure. In particular, at each time step of a FEM simulation a linearization point is added to the ROM if the reduced order state predicted by the ROM diverges too significantly from the FEM result. In other words, the ROM is built incrementally over the course of the simulation. 

For good ROM accuracy, the FEM simulation that is used for POD snapshot collection and linearization point selection should sufficiently cover the full range of possible robot motions. To ensure sufficient data is collected, a simple yet effective approach is to apply an open-loop control sequence in the FEM simulation that approximately spans the range of possible actuations for the robot. Specifically, we choose this sequence through Latin hypercube sampling of the soft robot's admissible actuations. A summary of the automated procedure for developing the piecewise affine ROM is given in Algorithm \ref{alg:rom_computation}, where \textbf{FEM} simulates the FEM model \eqref{eq:fom_dynamics} over one time step, \textbf{POD} computes the reduced order basis $U$, \textbf{project} corresponds to the projection \eqref{eq:project}, and \textbf{ROM} corresponds to simulating the piecewise affine ROM \eqref{eq:dynamics_rom}.
\vspace{-10pt}
\begin{algorithm}
    \caption{Defining Piecewise Affine ROM (Offline)}
    \label{alg:rom_computation}
    \begin{algorithmic}[1] 
        \Procedure{DefineROM}{$T_\text{sim}$, $r$, $W_q$, $W_v$, $\eta$}
            \State $\{u_k\}_{k=0}^{T_\text{sim}} \leftarrow $ \textbf{LatinHypercubeSample}($T_\text{sim}$)
            \State $\mathcal{X}^f = \{(q^f_{0}, v^f_{0}, 0)\}$
            \For{$k=\{0, \ldots, T_\text{sim}\}$}
                \State $(q^f_{k+1}, v^f_{k+1}) \leftarrow$ \textbf{FEM}($u_k$, $q^f_k$, $v^f_k$)
                \State $\mathcal{X}^f \leftarrow \mathcal{X}^f \cup (q^f_{k+1}, v^f_{k+1}, v^f_{k+1} - v^f_{k})$
            \EndFor
            \State $U \leftarrow$ \textbf{POD}($\mathcal{X}^f$, $r$)
            \State $\mathcal{P} = \{(q^f_{0}, v^f_{0})\}$
            \For{$k=\{0, \ldots, T_\text{sim}\}$}
                \State $(q^f_{k+1}, v^f_{k+1}) \leftarrow$ \textbf{FEM}($u_k$, $q^f_k$, $v^f_k$)
                \State $(q_{k+1}, v_{k+1})  \leftarrow$ \textbf{project}($q^f_{k+1}, v^f_{k+1}$)
                \State $(\tilde{q}_{k+1}, \tilde{v}_{k+1}) \leftarrow$ \textbf{ROM}($u_k$, $q_k$, $v_k$, $\mathcal{P}$)
                \If{$\lVert q_{k+1} - \tilde{q}_{k+1} \rVert_{W_q} + \lVert v_{k+1} - \tilde{v}_{k+1} \rVert_{W_v}  > \eta$}                
                    \State $\mathcal{P} \leftarrow \mathcal{P} \cup (q^f_{k}, v^f_{k})$
                \EndIf
            \EndFor
        \Return $U$, $\mathcal{P}$
        \EndProcedure
    \end{algorithmic}
\end{algorithm}
\vspace{-10pt}

\section{Reduced Order Optimal Control}\label{sec:control_scheme}
We now leverage the ROM \eqref{eq:dynamics_rom} to formulate a reduced order OCP to approximately solve the original OCP \eqref{eq:ocp}.
An output feedback control scheme is then defined which consists of three components: (1) the reduced order OCP, (2) a reduced order state estimator, and (3) a reduced order feedback control law.
In this scheme, the reduced order OCP optimizes a reduced order trajectory that the soft robot should follow and the state estimator incorporates measurements to provide an estimate of the robot's current (reduced order) state. The feedback control law then uses the state estimate to drive the robot to track the optimized trajectory.

\subsection{Reduced Order Optimal Control Problem}\label{sec:roocp}
A discretized optimal reduced order trajectory $(\mathbf{x}^*, \mathbf{u}^*)_k = (\{x^*_i\}_{i=k}^{k+T}, \{u^*_i\}_{i=k}^{k+T-1})$ for the robot is defined over a finite horizon $T$ by solving a reduced order approximation of the high-dimensional OCP \eqref{eq:ocp}:
\begin{equation}\label{eq:ocp_rom}
    \begin{split}
        \underset{x,{u}}{\text{minimize}} \quad &\lVert \delta z_{k+T} \rVert_{Q_f}^2 + \sum\limits_{j=k}^{k+T-1} \lVert \delta u_j \rVert_{R}^2 + \lVert \delta z_j \rVert_{Q}^2, \\
        \text{subject to}\quad &         x_{i+1} = g(x_i, u_i), \\
        & x_k = x_{0,k}, \\
        & u_i \in \mathcal{U}, \quad z_i \in \mathcal{Z}, \quad z_i = C_z x_i + z_{\text{ref}},
    \end{split}
\end{equation}
where $i=k, \ldots, k+T-1$, and $x_{0,k}$ is the initial state at time step $k$. This finite horizon problem is then solved in a receding horizon fashion to define the optimal trajectory over an arbitrarily long horizon. Specifically, the OCP is initialized at time $k=0$ by setting $x_{0,0} = \hat{x}_0$, where $\hat{x}_0$ is the reduced order state estimate. The OCP is then recursively solved every $T_r < T$ time steps (i.e. at time steps $k = T_r, 2T_r, \dots$ over the receding horizon $[k, k+T]$) by setting $x_{0,k} = x^*_{k}$, where $x^*_{k}$ is the optimal state from the previous solution (computed at time step $k-T_r$).

We solve the nonconvex OCP \eqref{eq:ocp_rom} using sequential convex programming (SCP), by solving a sequence of quadratic program (QP) approximations until convergence. Specifically, we use a slightly modified version of \cite{BonalliCauligiEtAl2019}.
Crucially, since the ROM is constructed such that $n \ll n^f$ this approach can enable real-time control. 

\subsection{Reduced Order Controller and State Estimator}\label{sec:control+state_est}
The reduced order OCP \eqref{eq:ocp_rom} defines an optimized \textit{open-loop} reduced order trajectory that the robot should follow. A simple output feedback control scheme is now defined to drive the robot to track this trajectory. 

To estimate the robot's current reduced order state from measurements a reduced order state estimator is defined as:
\begin{equation}\label{eq:state_estimator}
    \begin{split}
        \hat{x}_{k} =& g(\hat{x}_{k-1}, u_{k-1}) + L_k(y_k - C_y g(\hat{x}_{k-1}, u_{k-1})),
    \end{split}
\end{equation}
where $L_k \in \mathbb{R}^{n \times p}$ is the estimator gain, $y_k$ is the robot measurement, and $u_{k-1}$ is the previous control input. We choose the gain $L_k$ to be the extended Kalman filter gain based on the discrete-time ROM dynamics \eqref{eq:discrete_rom}.

The control applied to the robot is then computed using a reduced order linear feedback control law:
\begin{equation}\label{eq:feedback_law}
    \begin{split}
        u_{k} = u_{k}^* + G_{k}(\hat{x}_k - x_k^*),
    \end{split}
\end{equation}
where $G_{k} \in \mathbb{R}^{m \times n}$ is the controller gain matrix, and $(x^*_k,u^*_k)$ is the optimized state-input pair computed by the OCP \eqref{eq:ocp_rom}. 
To define the controller gains $G_{k}$ we propose a simple and computationally efficient method based on gain scheduling:
\begin{equation}\label{eq:feedback_gain}
\begin{split}
G_{k} = &\begin{cases}
G_{i}, \quad i = \arg\min_j \lVert x^*_k - x_j \rVert_W,
\end{cases}
\end{split}
\end{equation}
where the gains $G_i$ are the discrete-time LQR gains at each linearization point $(q_i, v_i)$, which can be computed \textit{a priori}. In particular, the gains $G_i$ are computed using the discrete-time reduced order dynamics matrices $A_{i,d}$ and $B_{i,d}$, and positive semi-definite cost matrix $Q_G \in \mathbb{R}^{n \times n}$ and positive definite control cost matrix $R_G \in \mathbb{R}^{m \times m}$ (e.g. $Q_G = C_z^T Q C_z$ and $R_G = R$).

The formulation of the proposed control scheme has a couple of practical advantages. First, since the OCP \eqref{eq:ocp_rom} defines the initial condition based on the previous solution the next solve can be started as soon as the previous solve is completed. For example, suppose the optimized trajectory $(\mathbf{x}^*, \mathbf{u}^*)_k$ has already been computed at time step $k$, then the OCP problem associated with time step $k + T_r$ can be solved starting at time step $k$ since $x^*_{k + T_r}$ is already known. If the model is discretized with a sampling time of $h$ this gives $hT_r$ seconds to solve the next OCP. Additionally, this choice of initialization can help with warm-starting of the SCP algorithm.
Second, even though the optimal trajectory computed by the OCP is discretized with a sampling time $h$, the state estimator and control law can be operated at a higher frequency by simply interpolating the optimal trajectory $(\mathbf{x}^*, \mathbf{u}^*)_k$.

\section{Simulation Results}\label{sec:results}
We now compare our method against two alternative approaches. First, we use a reduced order model predictive control (ROMPC) scheme based on a \textit{linearized} FEM model \cite{LorenzettiPavone2020b} to demonstrate the significant benefits of using a nonlinear model. 
Second, we demonstrate that our method performs comparably to a data-driven Koopman operator-based control scheme \cite{BruderGillespieEtAl2019} without suffering common drawbacks of data-driven methods, such as a loss in generality from building problem or task-specific models. These comparisons are discussed in more detail in Section \ref{subsec:results_comparison}.

\begin{figure*}[t]
    \centering
    \includegraphics[width=1\textwidth]{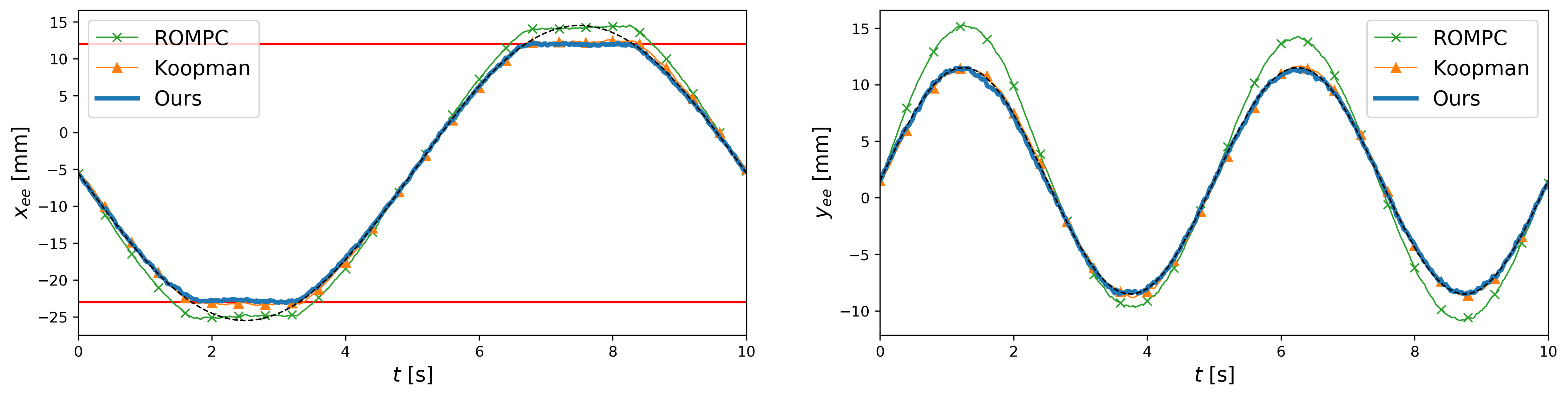}
    \caption{FEM model simulation results for trajectory tracking with the Diamond soft robot. The green line represents the ROMPC controller \cite{LorenzettiPavone2020b}, orange is the Koopman controller \cite{BruderGillespieEtAl2019}, blue is our proposed control scheme. The red lines represent constraints, and the black dashed line indicates the desired trajectory (which is partially infeasible due to the constraints). These results demonstrate the ability of our proposed controller to allow the soft robot to perform dynamic tasks such as constrained trajectory tracking.}
    \label{fig:diamond_results_time}
\end{figure*}

We compare these approaches in simulation using the elastomer ``Diamond'' soft robot shown in Figure \ref{fig:robots_fem}. This robot is fixed at the base and is actuated by controlling the tension in four cables that are attached at the robot's ``elbows''. The ``top'' of the robot will be referred to as the ``end effector''.
We use the open-source SOFA framework \cite{FaureDuriezEtAl2012,CoevoetMorales-BiezeEtAl2017} for finite element model-based simulations, and the mesh used in our experiments can be found in the SOFA Soft Robots plugin\footnote{\href{https://github.com/SofaDefrost/SoftRobots}{github.com/SofaDefrost/SoftRobots}}.
The FEM model used to simulate the Diamond robot consists of $\mathcal{N} = 1628$ nodes (i.e. $n^f = 9768$), the nonlinear internal forces $F^f(q^f, v^f)$ are defined by a linear stress-strain law, and gravity is assumed to be the only external force $P^f$.
The measurement model includes the position and velocity of the end effector and the four ``elbows'' of the robot (i.e. $y \in \mathbb{R}^{30}$), and additive Gaussian measurement noise is included in the FEM simulation.
We consider a control application where the performance variable is the position of the robot's end effector (i.e. $z = [x_{ee}, y_{ee}, z_{ee}]^T$). In particular, a constrained optimal control problem is formulated to drive the end effector position $(x_{ee}, y_{ee})$ to track a desired trajectory $(x_{d,ee}, y_{d,ee})$ subject to position constraints, as shown in Figures \ref{fig:diamond_results_time} and \ref{fig:diamond_results_infinity}. Note that this type of dynamic control problem cannot be addressed with kinematics-based controllers, and the addition of constraints precludes the use of many data-driven control methods.

\subsection{Controllers}\label{sec:controllers_comparison}
For the proposed approach we construct a piecewise affine ROM for the Diamond robot using the procedure in Algorithm \ref{alg:rom_computation}. In particular, we use acceleration snapshots in the POD step to define a reduced order subspace with dimension $r = 21$ (i.e. $n = 42$), we choose $\fomref{v} = 0$, and select $\fomref{q}$ to be the equilibrium configuration of the robot with no actuation. For selecting the linearization points we choose the threshold parameter $\eta$ (with $W_q = 0, W_v = I$) such that a total of $N = 642$ points $(q_i,v_i)$ are selected. Additionally, the piecewise affine ROM is defined with $W=\text{blkdiag}(0,I)$, such that only the robot's configuration $q$ is used to select the nearest linearization point. The reduced order OCP \eqref{eq:ocp_rom} is then defined with a horizon of $T=5$, a rollout horizon of $T_r = 2$, and the ROM is discretized with a sampling time of $h=0.05$ seconds. 
By interpolating the optimal trajectory computed by the OCP, the feedback controller \eqref{eq:feedback_law} and state estimator \eqref{eq:state_estimator} operate with a sampling time of $0.01$ seconds.

The ROMPC controller \cite{LorenzettiPavone2020b} uses a ROM generated by linearizing the FEM model \eqref{eq:fom_dynamics} around the point $\fomref{x}$ and then reusing the POD projection from our proposed controller. We consider the same horizons $T, T_r$ and sampling time $h$ as in our controller, and the ROMPC feedback controller and state estimator also operate with a sampling time of $0.01$ seconds by interpolating the optimized trajectory.
Finally, for the Koopman operator-based controller \cite{BruderGillespieEtAl2019} we build a model for the performance variables $z$ with a delay $d=1$ and all monomials of maximum degree $2$ to define a lifted state of dimension $N_K=66$ and a sampling time of $h = 0.05$ seconds. 
The data used to build this model came from the same FEM model simulation used to define our proposed controller.

\subsection{Results}\label{subsec:results_comparison}
\begin{figure}[t]
    \centering
    \includegraphics[width=0.90\columnwidth]{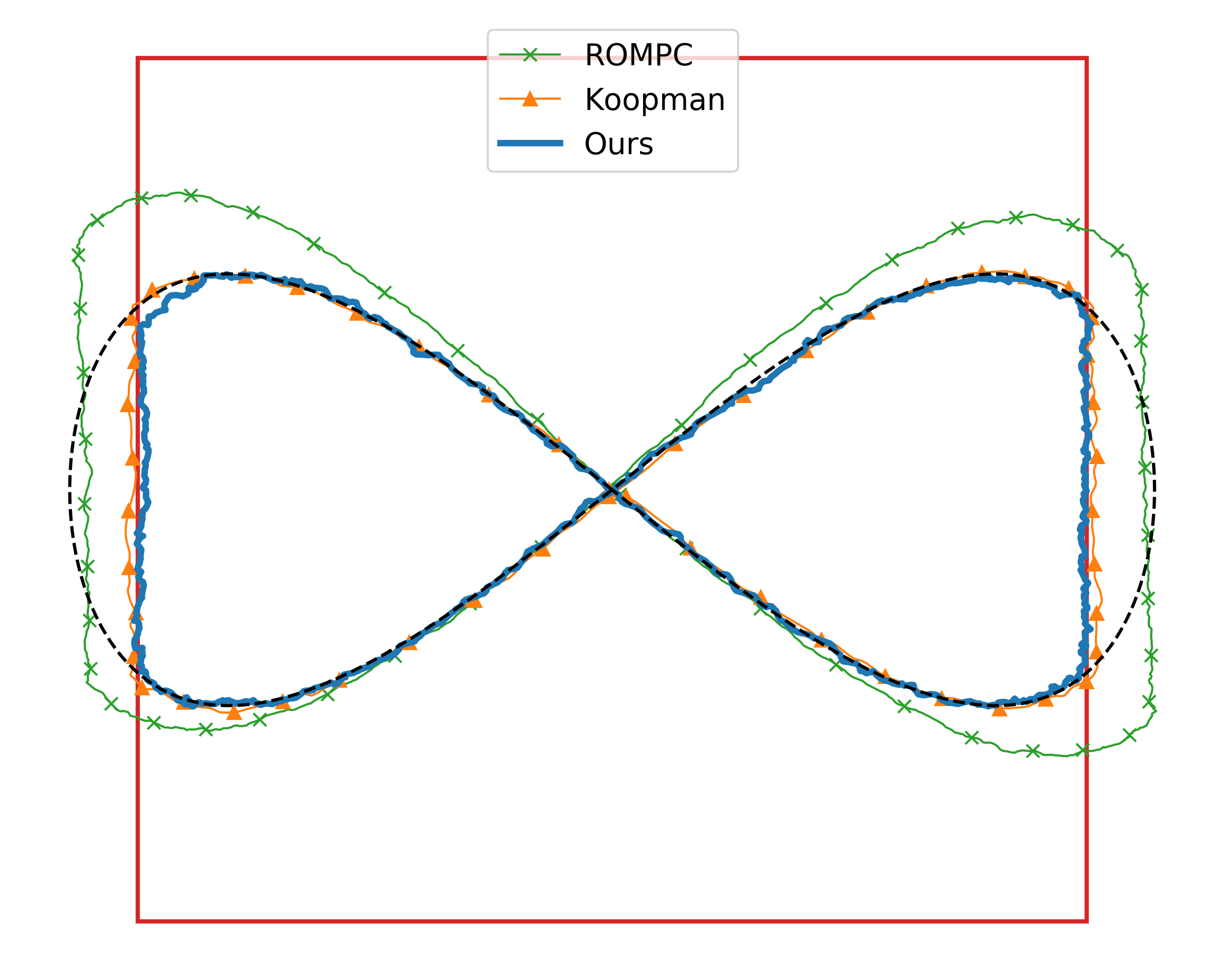}
    \caption{FEM model simulation results for the Diamond soft robot with the ROMPC \cite{LorenzettiPavone2020b}, Koopman \cite{BruderGillespieEtAl2019}, and our proposed controller. The black dashed line indicates the desired trajectory, and the interior of the red box is the admissible region defined by the constraints. Note that in addition to the advantages discussed in Section \ref{subsec:results_comparison}, our proposed approach performs comparably to the Koopman approach with respect to tracking and constraint satisfaction. The poor performance of ROMPC motivates the need for nonlinear model-based controllers.}
    \label{fig:diamond_results_infinity}
\end{figure}

Simulation results of the Diamond FEM model are presented for each controller in Figures \ref{fig:diamond_results_time} and \ref{fig:diamond_results_infinity}. As can be seen, the ROMPC scheme, which uses a \textit{linearized} ROM, offers poor tracking and severely violates the constraints. In contrast, our approach and the Koopman controller offer good tracking performance and generally satisfy the constraints, highlighting the importance of capturing the soft robot's nonlinear behavior. 

\begin{table}[t]
\caption{\label{tab:comptimes} A comparison of the mean square error (MSE) and the time spent solving QPs in each controller using Gurobi \cite{GurobiOptimization2020}, where the reported value for our method considers the \textit{cumulative} sum of all QP solve times in the SCP algorithm (on a 2.5 GHz Intel Core i5 processor with 8GB of RAM). These results show our FEM-based optimal control scheme achieves state-of-the-art performance and is real-time capable.}
\maketitle
\centering
\setlength\tabcolsep{5pt}
\renewcommand{\arraystretch}{1.2}
\begin{tabular}{|c"c"c|c|c|}
\cline{1-5}
\multicolumn{1}{|r"}{} & Tracking Error & \multicolumn{3}{c|}{Computation} \\ \hline
\multicolumn{1}{|r"}{Method} & MSE (mm$^2$) & Mean (ms) & Min. (ms) & Max. (ms)  \\ \hline
\multicolumn{1}{|r"}{ROMPC} & 5.80 & 10 & 9 & 18  \\ \hline
\multicolumn{1}{|r"}{Koopman} & 0.17 & 21 & 17 & 38 \\ \hline
\multicolumn{1}{|r"}{Ours} & 0.07 & 17 & 10 & 34 \\ \hline
\end{tabular}
\end{table}
To demonstrate the computational requirements of each controller we report the amount of time spent solving QPs (which is the most significant computational component) in Table \ref{tab:comptimes}. Note that the ROMPC and Koopman controllers only solve one QP at a time while the proposed SCP approach may require multiple QPs to be solved. These results show that the proposed FEM model-based control scheme is real-time capable.

Our proposed approach offers similar performance to the Koopman operator-based controller in this simulation (see Table \ref{tab:comptimes}), while offering several advantages. First, the Koopman approach only models the behavior of a \textit{prespecified} choice of the performance variables $z$, while the FEM model in our approach captures the robot's entire state (independent of the choice of $z$) and can therefore be used for any number of different control problems. In contrast to our approach, the Koopman controller is also restricted to only consider outputs $z$ that are a subset of the measured variables $y$. Second, the dimension of the Koopman model does not scale well with the number of measured variables. For example, including all the measured variables $y$ (i.e. setting $z=y \in \mathbb{R}^{30}$) would result in a model of dimension $N_K=2145$ (using the same delay $d=1$ and all monomials with maximum degree $2$). Third, the Koopman controller must operate at whatever frequency the model is discretized at (and the frequency the QP is solved at), while our controller can be operated at much higher frequencies by subsampling the optimized trajectory.

\section{Conclusion}\label{sec:conclusion}
In this work we propose a novel approach for model-based optimal control of soft robots based on high-fidelity nonlinear finite element models. Notably, computational efficiency is achieved by defining a reduced order piecewise affine model to approximate the high-dimensional FEM model. The proposed controller enables output feedback constrained optimal control problems to be addressed, including setpoint and trajectory tracking problems. Simulation results are used to demonstrate the performance of the proposed approach in comparison to a state-of-the-art data-driven and linear FEM model-based constrained control method.

{\em Future Work:}
Future work includes validation of the proposed approach through hardware experiments as well as several extensions, including the use of more advanced nonlinear model reduction techniques (e.g. the energy-conserving sampling and weighting (ECSW) method \cite{FarhatChapmanEtAl2015}), application to different types of soft robots (e.g. pneumatically actuated robots, which have a distributed actuation force), the ability to handle scenarios where the robot makes and breaks contact with the environment, and whether data-driven techniques could be used to augment the current approach. Additional theoretical analysis is also needed to study whether performance, stability, and constraint satisfaction guarantees can be made within the proposed framework.


\addtolength{\textheight}{-11cm}   

\bibliographystyle{IEEEtran}
\bibliography{./main,./ASL_papers}

\end{document}